\RequirePackage{fix-cm}
\documentclass[smallextended]{svjour3}       
\smartqed  
\usepackage{graphicx}
\usepackage{multirow}
\usepackage{longtable}
\usepackage{array}
\usepackage{arydshln}
\usepackage{subcaption}
\usepackage{appendix}
\usepackage{natbib}
\usepackage{hyperref}
\usepackage{xcolor}
\usepackage{url}

\journalname{Language Resources and Evaluation}

\begin{document}

\title{ArgRewrite V.2: an Annotated Argumentative Revisions Corpus\thanks{This material is based upon work supported by the National Science Foundation (NSF) under grant \#1735752.\\ \\
This preprint has not undergone post-submission improvements or corrections. The Version of Record of this article is published in Language Resources and Evaluation, and is available online at \texttt{\url{https://doi.org/10.1007/s10579-021-09567-z}}.
}
}
%

\subtitle{}


\author{Omid Kashefi\textsuperscript{1} \and
        Tazin Afrin\textsuperscript{1} \and \\
        Meghan Dale\textsuperscript{2} \and
        Christopher Olshefski\textsuperscript{2} \and
        Amanda Godley\textsuperscript{2} \and
        Diane Litman\textsuperscript{1,2} \and
        Rebecca Hwa\textsuperscript{1}
}

\institute{Omid Kashefi \\       
           \email{kashefi@cs.pitt.edu}\\
            \at \textsuperscript{1} School of Computing and Information, University of Pittsburgh 
           \at \textsuperscript{2} Learning Research and Development Center, University of Pittsburgh
}

\date{2022}

\maketitle

\begin{abstract}
Analyzing how humans revise their writings is an interesting research question, not only from an educational perspective but also in terms of artificial intelligence. Better understanding of this process could facilitate many NLP applications, from intelligent tutoring systems to supportive and collaborative writing environments.
Developing these applications, however, requires revision corpora, which are not widely available.
In this work, we present ArgRewrite V.2, a corpus of annotated argumentative revisions, collected from two cycles of revisions to argumentative essays about self-driving cars. 
Annotations are provided at different levels of purpose granularity (coarse and fine) and scope (sentential and subsentential). 
In addition, the corpus includes the revision goal given to each writer, essay scores, annotation verification, pre- and post-study surveys collected from participants as meta-data.
The variety of revision unit scope and purpose granularity levels in ArgRewrite, along with the inclusion of new types of meta-data, can make it a useful resource for research and applications that involve revision analysis. 
We demonstrate some potential applications of ArgRewrite V.2 in the development of automatic revision purpose predictors, as a training source and benchmark.

\keywords{Revision \and Subsentential Revision \and Revision Purpose \and Annotated Corpus  \and Argumentative Essay}
\end{abstract}

\newpage

\section{Introduction}
\label{intro}
Writing is an essential human activity for organizing and understanding complex ideas \citep{Westby2010SummarizingTexts}, and revising is an important part of that process. The process of adding, deleting, rearranging or modifying words, phrases, sentences or paragraphs in one's writing not only improves the writing, but can  support more complex thinking about the subject at hand \citep{Flower1981AWriting,Allal2004RevisionProcesses}. Revisions to both wording and the main ideas of an essay are important but in different ways. Revisions to wording are important for improving the fluency and correctness of a text, whereas revisions to the main ideas help writers rethink and refine their argument or purpose \citep{beason1993feedback}. Researchers working on revisions have shown, however, that inexperienced writers typically focus only on changes to wording, not on the organization and main ideas (i.e., content)  of an essay \citep{cho2010student}. Best practices in writing instruction, therefore, emphasize the importance of teacher and peer feedback to support effective content-level revisions \citep{magnifico2014reconsidering}.

Computational researchers have recently taken interest in writers' revision processes for both scientific reasons as well as practical ones.
Scientifically, modeling how humans learn to present complex ideas has long been an active research area in artificial intelligence. Practically, a natural language processing (NLP) system that can model the revising process has many applications from intelligent tutoring systems \citep{merrill1992effective,roscoe2013writing,jacovina2016intelligent} to supportive writing environments \citep{Zhang2016ArgRewrite:Writings}.  

Developing these applications requires revision corpora, but only a limited set of them are available.
Some extant corpora have focused on Wikipedia revisions \citep{Daxenberger2012AArticles,Bronner2013UserHistories}; however, those revision properties and annotations are specifically designed for Wikipedia's collaborative writing environment, which hampers their applications to different and more general rewriting and revision analysis tasks. 
Another corpus of writing revisions is ArgRewrite V.1, which is a small collection of single-author college-level argumentative essays and their revisions, as well as a set of manually developed sentence-level annotations of revisions properties \citep{Zhang2017AWriting}.
This prior version of ArgRewrite took a first step toward the creation of a more general-purpose revision corpus. However, as a pilot study, the corpus development was limited in several ways: 
it did not study subsentential revision;
the annotation scheme did not make enough distinction between some revision purposes;
the students did not receive individualized feedback before they attempted to revise their drafts; and
the students' final drafts were not scored.

Therefore, in this paper, we present \textit{ArgRewrite V.2} corpus, which aims to alleviate these limitations in a number of ways:  
(1) revisions, which are in English, are annotated at both \textit{sentential} and \textit{subsentential} levels;
(2) the annotation scheme now includes a  \textit{precision} revision purpose for changes to the specificity of the sentences;
(3) a broader range of annotators with more rigorous training have coded the revisions with their argumentative purpose;
(4) the corpus is almost twice as big as its predecessor, now including 258 drafts (3 drafts from each of the 86 participants) with around 3.3K sentential and 2.5K subsentential revisions;
(5) the corpus comprises the \textit{personalized feedback} given to each student and the \textit{scores} for each draft.

Given the inclusion of new types of meta-data,
ArgRewrite may facilitate broader range of writing-related research from automatic essay scoring \citep{Burstein2013TheSystem,Taghipour2016AScoring,Amorim2018AutomatedRatings} to argumentative revision analysis  \citep{Connor1994PeerRevision,Zhang2015AnnotationRevisions,Afrin2020AnnotationWriting}. The relatively larger size and in-depth revision annotation of the corpus also makes it 
useful for supporting the development of application such as 
writing error detection and correction \citep{tetreault2010using,dahlmeier2011grammatical,Xue2014RedundancyWritings},
sentence simplification and compression \citep{vickrey2008sentence,coster2011learning,turner2005supervised,berg2011jointly,filippova2015sentence},
paraphrasing \citep{Kauchak2006ParaphrasingEvaluation,Barry2006CanPlagiarism,Berant2014SemanticParaphrasing},
precision and specificity detection \citep{Li2015FastSpecificity,Lugini2018PredictingDiscussion}. 
In this work, we demonstrate an application of ArgRewrite in developing automatic revision purpose classifiers and how its properties make an interesting case for studying classification improvement through data augmentation.

\section{ArgRewrite V.2: Essay Collection}
\label{sec:collection}

The design choices of a corpus will have significant impact on a corpus's usefulness and applicability. 
The guiding principle behind the design decisions of our revision corpus was to maintain a balance between the consideration of the inevitability of writing style idiosyncrasies and a focus on ubiquitous writing and revision phenomena that exist across writers. 
On the one hand, we intended to capture a wide variety of revision phenomena, expressed by a diverse population of writers; on the other hand, we needed to ensure that the revisions could be reliably annotated by trained domain experts and that the resulting annotations would be useful to the community for further analyses and application development.

In this section, we discuss the data collection methodology and the essay production process, while Section~\ref{sec:annotation} describes the annotation process; most of our discussion focuses on drafts (Draft1, Draft2, Draft3) and annotating revisions (Rev12, Rev23); the revisions are later used in an empirical NLP study. Exploring the rich auxiliary data -- including scores, expert feedback, and student questionnaires -- will be left to the follow-up studies.

\begin{figure}
    \centering
    
    \includegraphics[width=1\linewidth]{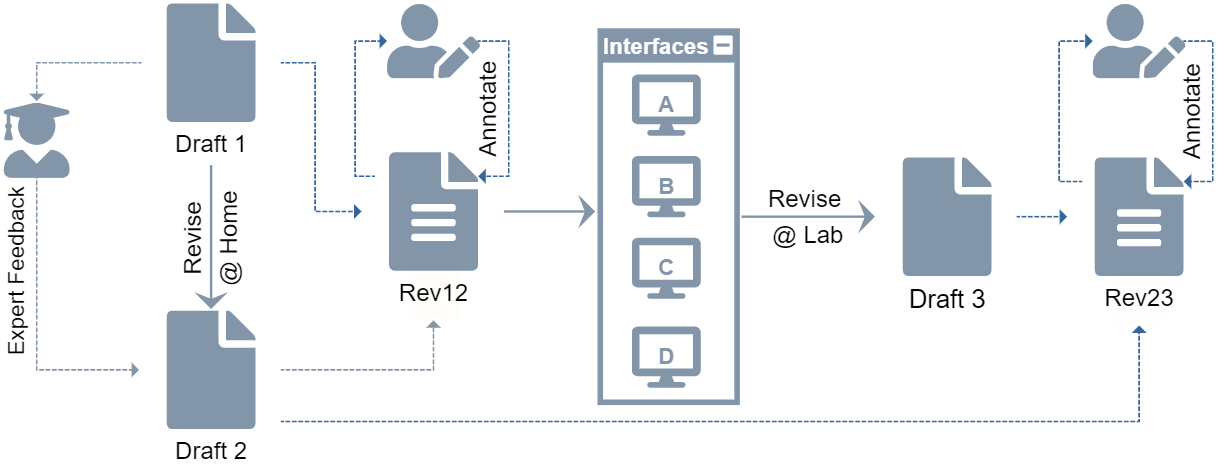}
    \caption{ArgRewrite V.2 Corpus Collection Process}
    \label{fig:proc}
\end{figure}

Figure~\ref{fig:proc}, illustrates an overview of the ArgRewrite V.2 corpus collection process:
a group of students produced essay drafts that were subsequently semi-automatically segmented and annotated by a group of experts. \textit{Draft1} is the initial version of the essay evaluated by an expert, resulting in a feedback text and a score, which is not shared with the writers. 
The second draft (\textit{Draft2}) was produced based on the draft and the feedback, . The first and the second draft were then semi-automatically segmented, aligned and annotated by the experts to produce the revision \textit{Rev12}.
This annotated revision was presented to the students via an interactive interface, in which the students prepared the third draft (\textit{Draft3}) under different interfaces. 
As before, the drafts Draft2 and Draft3 were aligned and annotated by the experts to produce the revision \textit{Rev23}.

\subsection{Collection Methodology}
\subsubsection{Participants}
\label{sec:participants}
Because we wanted to collect writing samples from individuals who were relatively familiar with the basic expectations of argumentative writing, we selected a university as the pool from which to collect our data \citep{BEACH1988TheWriting,CRAMMOND1998TheWriting}. 
Recruitment materials specified that participants must be aged 18 years old and older, either native English speaker or a non-native speaker possessing sufficient English proficiency (e.g., TOEFL score 100+).
The participant recruitment process was conducted through physical and electronic flyers posted throughout the University of Pittsburgh main campus and the Carnegie Mellon University. We were able to recruit 86 participants.
We expected that even within a pool of university students, we could recruit participants with a wide variety of argumentative writing skills (demographic information of participants are presented in Section~\ref{sec:stat:participants}). 


\subsubsection{Writing Task}
\label{sec:writing_task}
For the sake of consistency, all participants received the same instructions for a writing task which instructed them to develop an argument for or against self-driving cars that could serve as an op-ed piece in a local newspaper. 
In order to provide participants with comparable prior knowledge, each participant was provided with the same article about self-driving cars, organized according to the ``pros'' and ``cons'' of self-driving car technology. Participants were instructed to use the article to first summarize the advantages and disadvantages of self-driving cars before moving into their argument. They were advised that ``high quality'' op-ed pieces typically maintain ``a clear position on the issue'' and use ``supporting evidence'' as well as explanations of that evidence, and also include a ``counter-argument.'' They were told that such pieces also include clear organization, precise word choice and correct grammar. 
See Appendix~\ref{apx:prompt} for the prompt text.

In contrast to this task, our prior study did not provide a common reading for participants to cite. We believe the common reading materials served as a unifying force, making the argumentative essays more comparable, so that the corpus focus is more on revisions than the writers' prior knowledge. 

\subsection{Collection Process}
\label{sec:collection:process}

We collected \textit{three} drafts of argumentative essays from the participants in order to compare revision differences at different stages of rewriting. This process required each participant to take part in three sessions (refer to Figure~\ref{fig:proc}). In each session, participants were asked to write or revise their draft in approximately an hour. The sessions were organized as follows:

\paragraph{Draft1.} This session took place at home.
Participants were sent an email with a link to a pre-study questionnaire about their demographic background and self-reported writing background (see Appendix~\ref{apx:pre-study} for more details). 
Upon completion of the questionnaire, they were instructed to perform the writing task, described in Section~\ref{sec:writing_task}. 

\paragraph{Draft2.} This session also took place at home. 
After a few days, each participant received personalized formative feedback from a human expert on their first draft (e.g., ``Your essay’s sequence of ideas is inconsistent, with some clear and some unclear progression.'' See Appendix~\ref{apx:feedback:example} for a more detailed example of personalized feedback message\footnote{Personalized feedback is an augmentation over the previous version of the corpus, in which all participants received the same feedback (see Appendix~\ref{apx:feedback:v1}).}).
Feedback was provided via email and aligned with the writing criteria we later used to assess the quality of each draft (see Appendix~\ref{apx:feedback:rubric} for the scoring rubric). The feedback included 23 identified strengths and 23 weaknesses of the first draft. Participants were then asked to revise the first draft based on the feedback and resubmit the essay online.

\paragraph{Annotated Revisions I (Rev12).} To begin the annotation process, we first aligned Draft1 and Draft2 at sentence level; this was performed semi-manually, using a method that considered word-level similarity between sentences of different drafts and their ordering
\citep{Zhang2014Sentence-levelDetection}.
Then, revised sentences were automatically segmented into subsentential revision units, using a method that merged linguistically related sequences of word-level edits (add, delete, or modify) into a subsentential change  \citep{Xue2014ImprovedSentences}.
Finally, a trained annotator manually coded the perceived purpose of each revision unit (at the sentential and subsentential level), following the annotation guideline (see Appendix~\ref{apx:guideline}).
These annotations served as the ``Wizard of Oz'' feedback for the participants' next session. 

\paragraph{Draft3.} In this third and final session, participants were asked to view one of the four ArgRewrite web-based interfaces and then write and revise their third draft in a designated computer lab at the University of Pittsburgh.
Each participant was randomly assigned to one of the following four interfaces, which provided different types of feedback on how the participant had revised from Draft1 to Draft2\footnote{Further considerations about the interface design and user interactions are outside the scope of this paper; we discussed them separately elsewhere
\citep{Afrin2021EffectiveWriting}. Please refer to Appendix~\ref{apx:interfaces} for some screenshots of these interfaces.}.

\begin{itemize}
    \item \textit{Interface A:} only the sentences without any further feedback
    \item \textit{Interface B:} sentence-level differences, as a surface or content revision
    \item \textit{Interface C:} sentence-level differences with fine-grained revision purpose
    \item \textit{Interface D:} subsentential differences with fine-grained revision purposes
\end{itemize}

During the lab session, all participants, except those assigned to Interface A, were asked to agree or disagree with the annotator-recognized revision purposes shown by the system (i.e., Rev12). 
The annotation verification information could be used, for example, for analysing the impact of the difference between the the system’s recognized and the participant’s actual revision intents.
Then, all participants were asked to revise and submit their final draft and fill out a post-study questionnaire about their experiences (see Appendix~\ref{apx:post-study}).

\paragraph{Annotated Revisions II (Rev23).} After the participants submitted their Draft3, the revisions between Draft2 and Draft3 were coded by the trained annotator in the same process as annotating Rev12. Although our data collection stopped at Draft3, the participants' final round of revisions could also have been annotated and presented to the writers (via their preferred interface) to aid them with revising further drafts.

\subsection{Statistics}
Upon completion of collecting the essays, it is useful to review some corpus statistics; they help to assess whether the collected data matched the design goals we set out to achieve.
Here, we look into the participants' diversity and textual statistics of the collected essays.

\subsubsection{Participants}
\label{sec:stat:participants}
Table~\ref{tab:participants} shows the demographic statistics of the students who participated in the study.
Aiming to study with a diverse population of university students, we ended up recruiting a mixture of undergraduate students (58\%), graduate students (28\%), and some non-students, mostly post-docs and lecturers (14\%), where 80\% were native and 20\% were non-native English speakers, including 6 Chinese, 4 Hindi, 1 Vietnamese, 1 Tulu, 1 Telugu, 1 Japanese, 1 Korean, 1 Turkish, and 1 Kazakh native speakers. 

\begin{table}[h]
    \centering
    \caption{Participants' Demographic Statistics}
    \begin{tabular}{|r|cc||c|}
        \hline
         \multirow{2}{*}{\bf Education Level}   & \multicolumn{2}{c||}{\bf Language Proficiency} & \multirow{2}{*}{\bf Overall}  \\    
         \cline{2-3}
                                & { Native}  & { Non-Native} &  \\
        \hline \hline
        {Undergraduate}  & 46            & 13       & 50        \\
        {Graduate}       & 11            & ~4       & 24        \\
        {Other}          & 12            & ~0       & 12        \\
        \hline \hline
        {\bf Overall}           & 69            & 17       & 86 \\
        \hline 
    \end{tabular}
    \label{tab:participants}
\end{table}

\subsubsection{Essays}
Table~\ref{tab:textual_statistics} shows the textual statistics of the collected essays, including the number of essays, paragraphs, sentences, and words.
The corpus includes 258 essays, collected through 2 cycles of revisions from the participants.
In addition to having more essays in ArgRewrite V.2, an average of 29 sentences and 582 words per essay indicates that the essays are also much larger compared to the essays in the prior version of the corpus (53\% more sentences and 30\% more words per essay).
We also observe that, when participants proceed with their revisions, essays become lengthier -- as Draft2 has more words and sentences than Draft1 (on average, 29 sentences in Draft2 vs. 26 sentences in Draft1), and Draft3 more than Draft2 (on average, 33 sentences in Draft3 vs. 29 sentences in Draft2).

\begin{table}[h]
    \centering
    \caption{Textual Statistics of the ArgRewrite Corpus}
    \begin{tabular}{|r|cccc|}
        \hline
        {\bf Draft}     & {\bf Essays}  & {\bf Paragraphs} (avg)   & {\bf Sentences} (avg) & {\bf Words} (avg)  \\
        \hline  \hline
        {1}      & ~86            &   ~~405 (5)           & 2,216 (26)     & ~44,391 (516) \\
        {2}      & ~86            &   ~~451 (5)           & 2,461 (29)     & ~48,832 (568) \\
        {3}      & ~86            &   ~~488 (6)           & 2,814 (33)     & ~57,163 (665) \\
        \hline \hline
        {\bf Overall}   & 258            & 1,344 (5)           & 7,491 (29)     & 150,386 (582) \\
        \hline
    \end{tabular}
    \label{tab:textual_statistics}
\end{table}

\section{ArgRewrite V.2: Annotation}
\label{sec:annotation}
This section discusses our annotation scheme design, the annotation process itself, as well as some statistics of the annotated corpus.

\subsection{Annotation Scheme}
\label{sec:annotation:schema}
Our aim in developing a revision corpus is to understand why a writer makes certain revisions. Toward this end, we analyze the {\textit purposes} of edits -- are they primarily to improve readability or to convey different ideas? There are, however, many possible schemes to annotate these revisions. For example, we might opt to record fairly factual operations (text added, deleted, modified) or we might annotate the reasons for these operation, which may be more subjective. There is also the question of the appropriate scope of a revision; for example, if a relative clause is added to a noun, would that be considered an edit at the phrasal level, sentential level, and/or paragraph level? In developing the annotation scheme, we consider both the scope of the revision unit and the granularity of the purpose categories.


\subsubsection{Scope of the Revision Unit}
\label{sec:design:rev_unit}

In the prior version of the corpus, a revision was defined as an original \textit{sentence} paired with its revised version \citep{Zhang2017AWriting}. 
However, a sentence level revision can itself be a collection of multiple separate smaller revision units, which could have different revision purposes; little research has been done on the revision unit and scholars are uncertain whether a larger unit (e.g., at sentence or paragraph level) or a smaller unit (e.g., at phrase, or word, or character level) are more effective at supporting improvement in revision practices \citep{magnifico2014reconsidering}.
Therefore, for each revision, we decided to provide the annotations at both sentential and subsentential (phrase) levels to expand the corpus application to revision studies at both levels.
In the prior ArgRewrite corpus, semantically similar sentence pairs were aligned and annotated for one revision purpose, as a (sentential) revision. 
In the current version, however, we go further by segmenting sentential revisions into their subsentential revised units, which can be annotated independent of their corresponding sentential revisions.

\begin{figure}
    \includegraphics[trim=.45cm 9.5cm .2cm 1.5cm,width=1\linewidth]{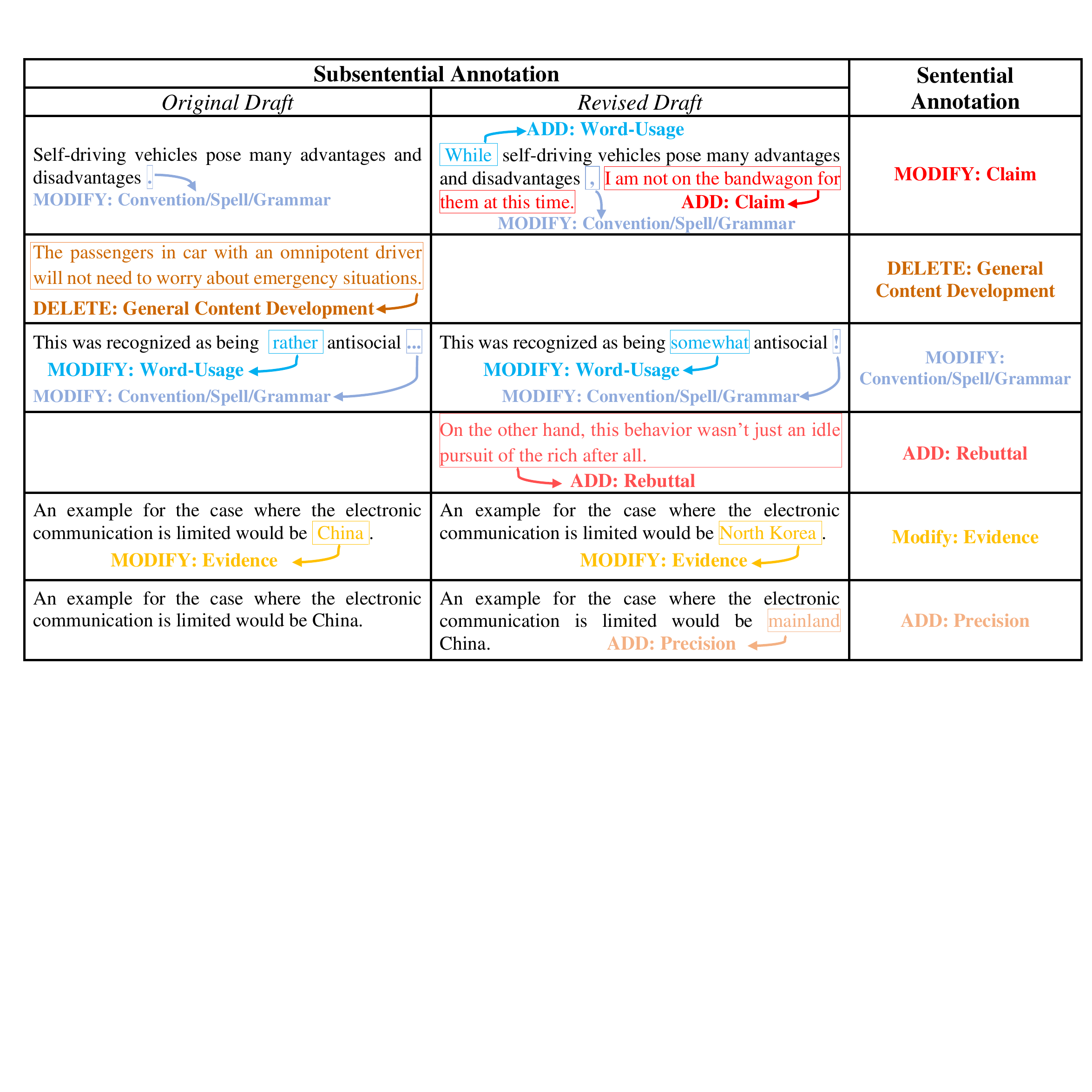}
    \caption{Example of Revisions with Sentential and Subsentential Annotations} 
    \label{fig:annotation-example}
\end{figure}

Figure~\ref{fig:annotation-example} shows some examples of revisions, annotated as both sentential and subsentential units.
As an instance, in the first sentence pair, ``While'' is labeled as \textit{ADD: Word-Usage} in the beginning of the sentence, the component ``I am not on the bandwagon. . . '' is labeled as \textit{ADD: Claim}, and the change of punctuation mark from period (``.'') in original sentence to comma (``,'') in the revised sentence is labeled as \textit{MODIFY: Convention/Spell/Grammar}.
Annotations at the sentential scope, however, would simply label the whole sentence pairs as \textit{MODIFY: Claim}. Thus, depending on the scope of revision units, annotations can vary.   

\subsubsection{Granularity of the Revision Purpose Categories}
\label{sec:design:rev_granularity}

Similar to our design choices for the scope of the revision units, we also annotated the revision purposes at multiple levels of granularity. 
We built upon the annotation schema developed for the prior version of the corpus \citep{Zhang2015AnnotationRevisions}, wherein revisions were annotated for their edit \textit{operation} and argumentative \textit{purpose}. 
In this version of the corpus, we have updated some of our definitions and included some new categories. 

Each change made to an essay was annotated with \textit{Add}, \textit{Delete}, or \textit{Modify} revision operations, according to how it related to its original version in the prior draft of the essay. 
These operations correspond to the addition or deletion of a whole sentence, or modification of an already existing sentence during the revision.
At the subsentential level, however, a modified sentence could be revised by adding a few phrases to it, or deleting or modifying some of its phrases, so may receive different annotations based on its substantial unit changes (see Figure \ref{fig:annotation-example}). 

This corpus provides annotations at two different grain sizes. At the coarser grain size, revision units were annotated as either \textit{Content} revisions (i.e., changes to main ideas of the essay) or \textit{Surface} revisions (changes to the grammar, usage, or word choice). 
At the finer grain size, revisions were annotated for three different subcategories of surface revisions: 
\textit{Word Usage} (WRD), 
\textit{Spelling and Grammar} (SPL), 
\textit{Organization} (ORG) revisions. 
Content revisions are further categorized into six subcategories: 
\textit{Claim} (CLM),
\textit{Evidence} (EVD),
\textit{Reasoning} (RSN), 
\textit{Rebuttal} (RBL),
\textit{General Content Development} (GCD) revisions, or \textit {Precision} (PRN). 

The latter label, \textit{precision}, is new in this corpus and refers to words that are edited to affect the specificity of the sentence. The purpose of such revisions is deemed to be at a content level, even though the writer may change only a few words such that the edit resembles a surface change. 
An example of such revisions is shown in the last row of Figure~\ref{fig:annotation-example}: the original sentence was revised by adding the word ``mainland'', which makes it more specific by excluding some special administrative regions from the original claim of the sentence.   
For more details on the annotation guideline, see Appendix~\ref{apx:guideline}.

\subsection{Annotation Process} 
\label{sec:annotation:process}
Compared to our previous study, we decided to recruit more domain experts to annotate the corpus: 
an expert in argumentative analysis; 
an expert in AI and education; 
as well as a computer scientist trained in argumentative analysis. 
Having a larger number of annotators allowed us to study a more comprehensive inter-annotator agreement and annotation quality assurance. 
The three domain expert annotators were trained based on the annotation guideline (see Appendix~\ref{apx:guideline}). 
During the training process, annotators coded 5 revised essays with sentence-level revisions and 2 revised essays for subsentential revisions from our prior corpus, then discussed their annotation intuitions and disagreements.
After the annotation training, we ran a pilot version of our new study to collect 5 revised essays, then each annotator coded these essays. 

Table~\ref{tbl:kappa} shows the inter-annotator agreement on coding the pilot study revisions with the coarse and fine-grained revision purposes at sentential and subsentential levels, calculated as Fleiss' kappa \citep{Fleiss1971MeasuringRaters}.
As expected, annotators had higher level of agreement in coding a coarser-grained category scheme (2 categories: surface or content) compared to a finer one (9 categories: detailed revision purposes).
They also agreed more on subsentential annotations than sentential annotations. 
One reason could be that a subsentential change is made to serve just one revision purpose, while a sentential revision might be an amalgamate of multiple smaller changes, each made for different argumentative purpose, so merging these different purposes into one clear revision purpose might be harder to distinguish each of them (see Figure~\ref{fig:annotation-example}).

Nevertheless, since the annotators were able to reach substantial agreement on both sentential and subsentential revisions, each \textit{Rev12} and \textit{Rev23} file from all interfaces was randomly assigned to the annotators, so each annotator coded about one third of the ArgRewrite corpus. 
Revisions from Interface B and Interface C were annotated with fine-grained categories at sentence-level and revisions from Interface A and Interface D were annotated with fine-grained categories at both sentence and subsentential levels. 

\begin{table}[h]
    \centering
    \caption{Inter-Annotators Agreement (Fleiss' Kappa)}
    \begin{tabular}{|r|cc|}
         \hline    
         \multirow{2}{*}{\bf Revision Unit} & \multicolumn{2}{c|}{\bf Category Granularity} \\
         \cline{2-3}
          & {Coarse} & {Fine}\\
         \hline \hline    
        {Sentential} &  .71   & .65  \\
        {Subsentential} &  .92   & .78  \\
         \hline    
    \end{tabular}
    \label{tbl:kappa}
\end{table}

\subsection{Statistics}
\label{sec:annotation:stat}

Collecting a wide variety of revision phenomena that are representative of the revision behaviour of the students is an important aspect of developing a revision corpus. 
Table~\ref{tab:rev_purpose} shows the distribution of annotated revision \textit{purposes} for \textit{sentential} and \textit{subsentential} revision unit.
The corpus contains 3,238 sentential (84\% more than prior version) and 2,596 subsentential (new in ArgRewrite V.2) revisions annotated with both fine and coarse revision purpose category labels. 
Also, although all essays in the corpus were annotated at the sentence level, only the essays of participants using Interfaces A and D include annotations at the sub-sentential level. 

\setlength\tabcolsep{1.55pt} 
\begin{table}[b]
    \centering
    \caption{Revision Purpose Statistics of Sentential and Subsentential Revisions}
    \begin{tabular}{|r|cc:c|cc:c|}
        \hline
        \multirow{2}{*}{\bf Purpose~~~}    & \multicolumn{3}{c|}{{\bf Sentential}}     & \multicolumn{3}{c|}{{\bf Subsentential}}  \\
        \cline{2-4} \cline{5-7}
                                        & { Rev12} & { Rev23} & {\bf Overall}  (avg) & { Rev12} & { Rev23} & {\bf Overall}  (avg)\\
        \hline \hline
        Word Usage      &   453   &   577   &   1,030 ~(6)    &   ~~445   &   ~~654   &   1,099 ~(13)\\
        Spell/Grammar             &   125   &   114   &   ~~239 ~~(1)   &   ~~137   &   ~~150   &   ~~267 ~~(3)\\
        Organization    &   ~52   &   ~25   &   ~~~77 ($<$1)  &   ~~~33   &   ~~~~2   &   ~~~35 ($<$1)\\
        \hdashline
        \textbf{Surface} &   630   &   716   &   1,346 (8)   & ~~615   &   ~~806   &   1,421 ~(17)\\
        \hline \hline

        Claim           &   154   &   ~80   &     ~~234 ~~(1)   &   ~~~89   &   ~~~44   &   ~~133 ~~(2)\\
        Reasoning       &   262   &   352   &     ~~614 ~~(4)   &   ~~140   &   ~~243   &   ~~383 ~~(5)\\
        Evidence        &   112   &   ~88   &     ~~200 ~~(1)   &   ~~~46   &   ~~~51   &   ~~~97 ~~(1)\\
        Rebuttal        &   ~22   &   ~20   &     ~~~42 ($<$1)  &   ~~~15   &   ~~~12   &   ~~~27 ($<$1)\\
        Precision       &   ~50   &   ~35   &     ~~~85 ($<$1)  &   ~~~88   &   ~~~59   &   ~~147 ~~(2)\\
        GCD           &   397   &   320   &     ~~717 ~~(4)   &   ~~183   &   ~~205   &   ~~388 ~~(5)\\
        \hdashline
        \textbf{Content}  &   997   &   895   &   1,892 ~(11)    &   ~~561   &   ~~614   &  1,175 (14)\\
        \hline \hline

        \textbf{Overall} (avg)&   1,627 (19)  &   1,892 (22)  &   3,238 (19)    &   1,176 (28)   &   1,420 (34)  &   2,596 (31)\\       
        \hline
    \end{tabular}
    \label{tab:rev_purpose}

\end{table}
\setlength\tabcolsep{6pt} 

As shown, the corpus includes a variety of surface and content revisions, however, some revisions such as choosing a better word to express an idea (word usage), provide reasoning for a claim (reasoning), or introducing general content to develop an argument (other), are more frequent, while changes to the organization of the essay, or rebutting an idea or changing the specificity level of the essay (precision), rarely happen in students' revisions. 

Sentential revisions are more inclined toward content changes (on average, 11 content change per draft vs. 8 surface revisions per draft ), while it is quite the opposite for subsentential revision (on average, 14 content change per draft vs. 17 surface revisions), which may imply that a bigger content revision could be made through, or include, some smaller surface changes.
This fundamental difference between purposes of the same revisions at different unit scopes may validate that our decision to annotate revisions as both sentential and subsentential units can actually make it useful for applications analyzing different units of text, which may have different annotation requirements.    

In general, students made slightly more changes when revising their second draft (on average, 22 sentential or 34 subsentential revisions in Rev23 per draft) than revising their first draft (on average, 19 sentential or 28 subsentential revisions in Rev12 per draft).

Another way to look at the revisions is from the edit \textit{operation} perspective to see if a revision is made by adding or deleting some text, or modifying some part of the essays.
Table~\ref{tab:rev_op} shows the distribution of edit operations for sentential and subsentential revision units.
There is a small difference (54 sentences) between the number of revisions that are annotated with edit operation and those that are coded with revision purpose (Table~\ref{tab:rev_purpose}). 
Some of the revisions were annotated with more than one revision purpose, which is violating our annotation guideline, so we discarded them from the current version of the corpus.  

As shown in Table~\ref{tab:rev_op}, most of the revisions  involved modifying a previously written sentence (on average, modification of 10 sentences or 14 phrases of a draft), and deletion are the less popular operation for revising essays (on average, deletion of 3 sentences or 6 phrases from a draft).

From the revision operation perspective, unlike the revision purpose annotations, different drafts were revised quite similarly at both sentential and subsentential scopes (on average, 19 edit operations at sentence-level and about 29 edit operations at phrasal-level in both Rev12 and Rev23).
This observation implies that different dimensions of annotation may express a different type of information and reveal different characteristics of the revision behaviour, therefore, including different type of annotations for revisions (operation, coarse-grained purpose, fine-grained purpose) can widen the usefulness of our corpus for a more diverse set of applications.

\setlength\tabcolsep{1.6pt} 
\begin{table}
    \centering
    \caption{Revision Operation Statistics of Sentential and Subsentential Revisions}
    \begin{tabular}{|r|cc:c|cc:c|}
        \hline
        \multirow{2}{*}{\bf Operation~~}    & \multicolumn{3}{c|}{{\bf Sentential}}     & \multicolumn{3}{c|}{{\bf Subsentential}}  \\
        \cline{2-4} \cline{5-7}
                                        & {Rev12} & {Rev23} & {\bf Overall}  (avg) & {Rev12} & {Rev23} & {\bf Overall}  (avg)\\
        \hline \hline
        Add         &   555   &   530   &   1,085 ~(6)   &   370   &   439   &   ~~809 (10)\\
        Delete      &   324   &   174   &   ~~498 ~(3)   &   245   &   243   &   ~~488 ~(6)\\
        Modify      &   777   &   932   &   1,709 (10)   &   568   &   580   &   1,148 (14)\\
        \hline \hline

        \textbf{Overall} (avg)&   1,656 (19)  &   1,636 (19)  &   3,292 (19)    &   1,183 (28)   &   1,262 (30)  &   2,445 (29)\\       
        \hline
    \end{tabular}
    \label{tab:rev_op}
\end{table}
\setlength\tabcolsep{6pt} 

\section{Corpus Availability}

The ArgRewrite V.2 is available 
from \texttt{\url{http://argrewrite.cs.pitt.edu}}
Participant identification information is anonymous and the corpus contains:

\begin{itemize}
    \item \textbf{Essays.} 258 raw text files of the written essays (86 of each draft).
    
    \item \textbf{Annotations.} 172 excel files: 86 Rev12 and 86 Rev23, grouped by the interface they are collected from.
    
    \item \textbf{Meta-Data.} The corpus is shipped with  students' responses to the pre-survey and post-survey questionnaires, and their annotation verification information.
    It also contains the score for each students' drafts (258 essays) and the expert feedback given to the Draft1 of the students (86 feedback).
    
\end{itemize}

\section{Example Usage: Revision Purpose Classification}
\label{sec:application}

The corpus will be useful for developing a variety of applications, from revision analysis to predicting whether a text chunk is expressing a general or specific piece of information. Additionally, the corpus affords the examination of a variety of feedback types and the types of revisions that follow. Scholars in composition and educational research might find it useful to map patterns in revisions to the four different  interfaces. Most readers of this paper, however, will likely be interested in the computational uses of the corpus, which we outline below. 
In this section, we demonstrate one example usage of the corpus -- the development of revision purpose classifiers, a component of an argumentative revision analysis system. 
The variety of revision unit scope and purpose granularity levels allows us to study a variety of revision classification tasks with different settings. 
Therefore, we experiment on a \textit{binary} classification task (Section~\ref{sec:binary}) and a \textit{multi-class} classification task (Section~\ref{sec:mlp}), both trying to predict the purpose of the sentential and subsentential revisions.

Since our main objective is to demonstrate the usefulness of our corpus for NLP applications, we do not develop highly domain specific features or complex models for the classification tasks.
Instead, we opt for some features (Section~\ref{sec:features}) and models (Section~\ref{sec:model}) that are widely applicable to many NLP applications \citep{Burstein2001TowardsEssays,Daxenberger2013AutomaticallyRevisions,Zhang2016UsingRevisions,Jabreel2018EiTAKATweets}.

\subsection{Features}
\label{sec:features}

We use a mixture of features to represent textual (length and position), syntactic (part-of-speech), semantic (embedding), and discourse (transition words) aspects of a revision as follow:

\begin{itemize}
    \item \textit{Length.} the length of the sentence in number of its words.\\
    
    \item \textit{Position.} the index (location) of the sentence in the essay's sentences.\\
    
    \item \textit{Embedding.} the vector representation of the sentence encoded using \textit{Universal Sentence Encoder} (USE), which is a pre-trained transformer-based encoder of greater-than-word length text \citep{Cer2018UniversalEncoder}.\\
    
    \item \textit{Part-Of-Speech.} the \textit{term frequency} representation of the sentence words' part-of-speech (POS) tags, predicted using \texttt{spaCy}\footnote{\texttt{https://spacy.io/}}. 
    See Appendix~\ref{apx:term-freq} for more details on how we generate the POS term frequency representation.\\
    
    \item \textit{Transition Words.} the term frequency representation of the transition words in the sentence. See Appendix~\ref{apx:trans-words} for the complete list of transition words we used to represent the discourse aspect of the revisions. 
\end{itemize}

Each \textit{sentential} revision is represented as the pair of \texttt{<old-sentence, new-sentence>}, which could be either between Draft1 and Draft2 (Rev12), or between Draft2 and Draft3 (Rev23). 
These sentences then transform into feature space.
Each \textit{subsentential} revision is represented as the pair of \texttt{<old-phrase, new-phrase>}, which could be either between Draft1 and Draft2 (Rev12), or between Draft2 and Draft3 (Rev23). 
To take the context of revised phrases into account, we extend the subsentential revision representation to include the sentences in which the phrases are used as: \texttt{<old-phrase\textbar old-sentence, new-phrase\textbar new-sentence>}.
Each context sentence and subsentential revision is transformed into feature space in the same process as for sentential revision representation, except for the \textit{position} feature of subsentential revisions, which is their starting index in the context sentence.
Note that in our experiments, we assume revisions are pre-segmented and pre-aligned at the desired revision scope level, based on the classification task settings.

\subsection{Training Settings}
\label{sec:model}
The choice of classifier model, tuning, datasets, and evaluation methodology of our experiments are as follow:

\begin{itemize}
    \item \textit{Classifier Model} 
    \begin{itemize}
        \item  \textit{XGB.} We opt to use XGBoost \citep{Chen2016XGBoost:System} as the learning algorithm in all classification tasks. 
        For each classification task, we explore a range of hyperparameter, including: \texttt{number of the estimators} $\in \{250, 500, 750, 1000 \}$, \texttt{maximum depth} $\in \{3, 4, 5\}$, and \texttt{learning rate} $\in \{.1, .05, .01 \}$. We pick the final setting through a randomized parameter search with cross-validation process\citep{Bergstra2012RandomOptimization}.

        \item \textit{Majority.} To better understand the classification result of each task and check whether we trained reasonable models, we compare the results with a simple majority classifier baseline, which assigns the most frequent revision purpose of the dataset to all revisions.\\
    \end{itemize}

    \item \textit{Sentential Dataset.} The sentential revision dataset contains 3,238 training examples collected from all four interfaces, with coarse and fine revision purpose annotation levels. We use this dataset to train the sentential binary and multi-class revision purpose classifiers.\\
    
    \item \textit{Subsentential Dataset.} The subsentential revision dataset contains 2,596 training examples collected from interface A and D, with coarse and fine revision purpose annotation levels.
    We use this dataset to train the subsentential binary and multi-class revision purpose classifiers.\\
    
    \item \textit{Evaluation.} Classifiers are evaluated in a 5-fold cross-validation process using \textit{average unweighted F-score} and \textit{Accuracy} measure.
\end{itemize}

\subsection{Binary Classification}
\label{sec:binary}
In this section, we experiment with the task of predicting whether the purpose behind a revision is to make a content-level change or a surface-level change.
Given that the ArgRewrite V.2 contains purpose annotation for sentential and subsentential annotations, we also investigate how coarse-grained revision purpose prediction tasks may differ for different revision scopes.
For the sentential classification, we trained the models on the \textit{sentential dataset}, and for the subsentential classification, we trained the models on the \textit{subsentential dataset}.
To better understand the contribution of different features (see Section~\ref{sec:features}), we experiment with three classification settings: (1) training only on semantic features of the revisions (referred to as \texttt{USE}), (2) training on textual, syntactic, and discourse features (referred to as \texttt{Features}), and (3) training using all of the features (referred to as \texttt{Features + USE}).
This ablation test can help to investigate the impact of semantics and how a pre-trained language model performs on the task.

Moreover, we study a cross-task experiment-- \textit{how does the model trained on sentential revisions performs on predicting the purpose of the subsentential revisions, and the other way around?} 
Since some features in the subsentential setting (e.g. positions, or the context sentence) are not applicable to the sentential setting, for this experiment, we use the model that is trained only on the embedding from USE, which is independent of tasks and domains.
To collect comparable results with other classification settings, we evaluate the model through 5-fold cross-validation, where in each fold, the training set, and the test set are picked from the corresponding splits from the sentential dataset and subsentential dataset, respectively.

Table~\ref{tab:bin} shows the average unweighted F-score and the accuracy (ACC) of our binary classification experiments for different revision scopes and settings\footnote{Hyperparameters: \texttt{estimators} = $500$, \texttt{maximum depth} = $4$, and \texttt{learning rate} = $.05$}. 
In general, we can observe that while predicting if a revision is a content or a surface change, it is slightly easier at sentential-level than subsentential-level. Both supervised models can achieve a high classification performance, which outperforms the majority baseline by a big margin, in all classification settings.
The embedding-only classifiers (\texttt{USE}) perform slightly better than the features-only classifiers (\texttt{Features}), while the classifiers that are trained on both (\texttt{Features+USE}) significantly outperform the feature-only classifiers.
However, the difference between \texttt{USE} and \texttt{Features+USE} is not significant, which suggests the domain-independent approach of using only embeddings from a pre-trained language model might also produce comparable results on predicting whether a revision purpose is a content or surface change.
Additionally, feature-only classifiers also produce promising classification results, suggesting this problem can also be addressed by more traditional solutions without sacrificing classification performance.  

As we expected the classification performance drops in cross-task evaluation, however, results also imply that the model trained on the sentential revisions could be used to predict the coarse-grained revision purpose for the subsentential revision with acceptable performance (F1: .71), while the subsentential model performs only as good as a majority classifier on predicting the purpose of the sentential revision.      
One possible reason for this could be that the subsentential revisions that are labeled as content changes are relatively longer than surface changes (on average, 9 words compared to 5 words). Thus, this model will predict a content label for (almost) all of the sentential revisions, which are longer than an average subsentential revision.  

\begin{table}
    \centering
    \caption{The F-Score and Accuracy of Binary Revision Purpose Classification, $^\dagger$indicates significantly better than \texttt{Features} ($p<0.05$).}
    \begin{tabular}{|c|r|cc||cc|}
        \hline 
                    \textbf{Scope} & \textbf{Model} & \textbf{Surface}   & \textbf{Content} & \textbf{AVG} & \textbf{ACC}\\
        \hline \hline
        \multirow{2}{*}{\bf \rotatebox{90}{\parbox{1.5cm}{\bf Sentence}}}
                                    & {Majority}    & .00       &  .73       & .37      & .58       \\
         \cdashline{2-6}
                                    & {Features}    & .89       &  .91       & .90      & .90\\
                                    & {USE}         & .91       &  .92       & .92      & .92\\
                                    & {Features + USE}         & {\bf.92}  &  {\bf .94} & {\bf.93} & {\bf.93}$^\dagger$\\
         \cline{2-6}
                                    & {Cross-Task (USE)}  & .70       &  .72       & .71      & .71\\
        \hline \hline

        \multirow{2}{*}{\bf \rotatebox{90}{\parbox{1.45cm}{\bf  Subsent.}}}
                                    & {Majority}    & .76       &  .00      & .38    & .61       \\
         \cdashline{2-6}
                                    & {Features}    & .88       &  .86      & .87    & .87\\
                                    & {USE}         & .89       &  .88      & .88    & .89\\
                                    & {Features + USE}         & {\bf.90}  &  {\bf.90} & {\bf.90} & {\bf.91}$^\dagger$\\
         \cline{2-6}
                                    & {Cross-Task (USE)}  & .09       &  .76    & .42      & .62\\
        \hline

    \end{tabular}
    \label{tab:bin}
\end{table}

\subsection{Multiclass Classification}
\label{sec:mlp}
In this section, we experiment with the task of predicting the fine-grained purpose of a revision and investigate how it may be influenced by the scope of the revision.
Similar to our binary classification experiment, for the sentential classification we train the models on the \textit{sentential dataset}, and for the subsentential classification we trained the models on the \textit{subsentential dataset}, but this time with fine-grained revision purposes as the supervision.

We also perform an ablation test to investigate the contribution of different types of features to the task, and
a cross-task experiment to see how does the pre-trained fine-grained revision purpose classifiers perform on predicting a purpose for revisions with different scope levels. 
For the same reasons mentioned in our binary cross-task experiment (see Section~\ref{sec:binary}), here we also use the models that are trained only on the embeddings.
To collect comparable results with other classification settings, we evaluate the model through 5-fold cross-validation, where in each fold, the training set and test set are picked from the corresponding splits from the sentential dataset and subsentential dataset, respectively.

Table~\ref{tab:cls2} shows the detailed average unweighted F-score and accuracy (ACC) of our fine-grained revision classification experiments for different revision scopes and settings\footnote{Hyperparameters: \texttt{estimators} = $750$, \texttt{maximum depth} = $5$, and \texttt{learning rate} = $.05$}. 
Intuitively, the multi-class classification is harder than the corresponding binary classification task, and here we also observe the multi-class classification experiments yield lower revision purpose prediction results than the binary classification experiments.
However, in contrast to our findings in binary classification experiments, we observe that predicting fine-grained revision purpose yields higher results for classifying subsentential revisions compared to sentential revisions. 
This observation is counter-intuitive because the \textit{subsentential dataset} contains fewer training examples than the \textit{sentential dataset} (3.2K vs. 2.6K, see Section~\ref{sec:model}).
Referring back to the inter-annotator agreements for annotating sentential and subsentential revisions, it seems that annotating revisions at the subsentential level is much easier than annotating them as sentential units.
A sentential level change might be the result of multiple subsentential changes, which do not necessarily have the same intended purposes, therefore, an amalgamation of different revision purposes uniting under their most prominent revision purpose (see Appendix~\ref{apx:guideline} for more details on annotation process).
As a result, classification models may find it harder to predict the purposes for sentential revisions, as opposed to subsentential revisions, which are atomic revisions with only one clear revision purpose, so are relatively easier to annotate and classify.

\setlength\tabcolsep{1.15pt} 
\begin{table}
    \centering
    \caption{The FScore and Accuracy of Fine-Grained Revision Purpose Classification, $^\dagger$indicates significantly better than \texttt{Features} ($p<0.05$), $^\ddagger$indicates significantly better than \texttt{Features+USE} ($p<0.05$)}
    \begin{tabular}{|c|r|ccc||cccccc||cc|}
        \hline
                     \multirow{2}{*}{\bf Scope} &  \multirow{2}{*}{\bf Model} & \multicolumn{3}{c||}{\bf Surface} & \multicolumn{6}{c||}{\bf Content} & \multirow{2}{*}{\bf AVG} &  \multirow{2}{*}{\bf ACC}\\
                    \cline{3-5} \cline{6-11}
                    & & {WRD} & {SPL} & {ORG} & {CLM} & {RSN} & {EVD} & {RBL} & {PRN} & {GCD} & & \\
        \hline \hline
        \multirow{2}{*}{\bf \rotatebox{90}{\parbox{1.62cm}{\bf Sentence}}}
        & {Majority}     & .45 & .00 & .00 & .00 & .00 & .00 & .00 & .00 & .00 & .05 & .29\\
         \cdashline{2-13}
        & {Features}     & .79 & .34 & .32 & .25 & .48 & .42 & .35 & .45 & .56 & .44 & .58\\
        & {USE}          & .78 & .38 & .35 & .33 & .58 & .45 & .37 & .54 & .59 & .49 & .62\\
        & {Features+USE}          & .79 & .39 & .35 & .37 & .60 & .48 & .37 & .54 & .60 & .51 & .63$^\dagger$\\
         \cdashline{2-13}
        & {+DA}          & .78 & .38 & {\bf .47} & {\bf .44} & .60 & {\bf .57} & {\bf .56} & {\bf .66} & .59 & {\bf .56} & {\bf .68$^\ddagger$}\\
         \cline{2-13}
        & {Cross-Task (USE)}   & .61 & .00 & .57 & .73 & .71 & .77 & .67 & .00 & .45 & .45 & .55\\
        \hline \hline
        \multirow{2}{*}{\bf \rotatebox{90}{\parbox{1.9cm}{\bf Subsentence}}}
        & {Majority}     & .61 & .00 & .00 & .00 & .00 & .00 & .00 & .00 & .00 & .06 & .44\\
         \cdashline{2-13}
        & {Features}    & .82 & .63 & .33 & .40 & .67 & .27 & .45 & .35 & .50 & .49 & .62\\
        & {USE}         & .83 & .65 & .33 & .48 & .60 & .46 & .45 & .35 & .46 & .54 & .66\\
        & {Features+USE}         & .83 & .71 & .33 & .44 & .67 & .46 & .45 & .45 & .56 & .57 & .70$^\dagger$\\
         \cdashline{2-13}
        & {+DA}         & .85 & .73 & {\bf .51} & {\bf .74} & .67 & {\bf .63} & {\bf .88} & {\bf .86} & .58 & {\bf .67} & {\bf .78$^\ddagger$}\\
         \cline{2-13}
        & {Cross-Task (USE)}   & .09 & .00 & .12 & .25 & .48 & .23 & .35 & .00 & .44 & .22 & .34\\
        \hline

    \end{tabular}
    \label{tab:cls2}
\end{table}
\setlength\tabcolsep{6pt} 

Similar to the binary classification experiments, we also observe that, while the \texttt{Features+USE} classifiers do not perform significantly better than the embedding-only classifiers, they significantly outperform the feature-only classifiers.
Moreover, the classification performance of the models drops when cross-evaluating them on predicting the purpose for different revision scopes.
Similar to our observations in binary classification experiment, 
the model trained on the sentential revisions performs better in predicting fine-grained revision purpose for the substantial revisions than the other way around.
This is in accordance with our intuition about the difference between the length of subsentential content and surface revisions, which may cause the subsentential models to predict a content-level purpose for (almost) all sentential revisions.

\subsubsection{Data Augmentation}
\label{sec:data-augmentation}

In the binary classification problem, the training examples are either surface or content revisions, so each of them comprises a reasonable amount of training examples, however, in our multi-class classification problem, training examples are distributed into nine classes, so compared to the binary case, we have fewer training examples for each class, while some, may seriously lack training examples (e.g., there are only 42 and 85 sentential training examples for the rebuttal and precision class, respectively). 
This training examples scarcity could be the main cause of the relatively lower prediction accuracy of fine-grained revision purposes, especially for under-represented revision purposes.
In order to investigate this, we study how data augmentation may help to improve the fine-grained revision purpose prediction performance by providing more training examples for under-represented classes.

We use a customized version of the synonym replacement (SR) augmentation strategy -- randomly pick a content word from the sentence and replace it with a synonym chosen at random \citep{Wei2019EDA:Tasks}, as our augmentation strategy to generate training examples. 
In general, we generate up to \textit{4} (on average: 3.4) augmented examples by substituting about 20\% of its content-words with their synonyms, which are retrieved using \texttt{sense2vec} contextual word embedding \citep{trask2015sense2vec}, for each examples of the underrepresented revision purposes, namely \textit{claims}, \textit{rebuttal}, \textit{evidence}, \textit{precision}, and \textit{organization}.
Our data augmentation strategy is discussed in detail elsewhere \citep{Kashefi2020QuantifyingAugmentation}.

During the cross-validation, each time, we expanded the training fold with new augmented examples and evaluate the model on the test fold. The rows indicated by \texttt{+DA} in Table~\ref{tab:cls2} show that incorporating data augmentation to generate more training examples can improve the fine-grained revision purpose classification at both sentential and subsentential levels.
Aside from overall classification improvements, we can also observe an average F-score improvement of around 30\% and 70\% for classifying the underrepresented sentential and subsentential revision purposes, respectively, when the training set is augmented with more samples for them.  
Therefore, with more training examples, the fine-grained purpose of revisions could also be precisely predicted. 

\section{Related Work}
\label{sec:relatedwork}

Early studies describe revision as a recursive process that involves both lexical and semantic changes~\citep{sommers1980,Fitzgerald1987ResearchWriting,Flower1981AWriting}. Those studies also show that effective writers' revision strategies differ from those of novice writers ~\citep{Flower1981AWriting}. Hence, more and more studies have focused on understanding students' revision efforts. However, research on writing revision is inadequate in NLP. Prior NLP research on writing has focused on analysis of a single drafts as opposed to multiple iterations of the same composition. Such studies have focused on, for example, esssay scoring ~\citep{attali2006b,Taghipour2016AScoring}, discourse structure analysis ~\citep{burstein2003m,falakmasir2014a} and
paraphrase detection ~\citep{Dolan2005AutomaticallyParaphrases,Barron-Cedeno2013PlagiarismDetection,Tan2014ACommunication,Vila2015RelationalCorpus}, grammatical or semantic error correction \citep{Dahlmeier2013BuildingEnglish,Yannakoudakis2011ATexts,Kashefi2018SemanticDetection}. The most closely related work to ours that has focused on revision are the bodies of literature on Wikipedia user edits or student academic essay revision.

Most related to our work is the Wikipedia revision analysis and categorization ~\citep{Daxenberger2013AutomaticallyRevisions,Bronner2013UserHistories,sarkar-etal-2019-stre}.
Revision categorization of user edits from Wikipedia focus on both coarse-level~\citep{Bronner2013UserHistories} and fine-grained~\citep{Daxenberger2012AArticles,Yang2017IdentifyingWikipedia,johnes2008wikirevision} categories. Although coarse-level categories (e.g., surface vs. content) can be generalized for academic writing, some fine-grained Wikipedia categories (e.g., vandalism) are specific to wiki scenarios. In academic writing, previous studies instead use fine-grained revision categories more suitable for student argumentative writing~\citep{toulmin_2003,Zhang2015AnnotationRevisions}.  The above studies focus on investigating the reliability of manually annotating and automatically classifying the revision categories. 
Other related works for categorizing revisions include measuring statement strength of revised sentences in academic writing~\citep{Tan2014ACommunication}, sentence-level revision improvement in argumentative writing~\citep{afrin2018improvement}, modeling revision requirement in wiki instructions~\citep{bhat-2020-wikihowinstructions}, etc.

There are many NLP-based writing assistant tools that were developed over the last few years. Such tools usually focus on grammar error correction of a single draft, few also provide high-level semantic error suggestions. 
For example, Grammarly~\citep{grammarly} provides feedback on grammatical error correction and fluency or word-usage,  ETS-writing-mentor~\citep{ets-writing-mentor} provides feedback to reflect on higher-level essay properties such as coherence, convincingness, etc. Other tools such as EliReview~\citep{elireview}, Turnitin~\citep{turnitin} are focused on peer feedback, plagiarism detection than focusing on student revision analysis. 
In contrast, our ArgRewrite revision assistant tool is focused on students' revision between two drafts of an essay. 
The prior version of our system provided feedback based on detailed revision categorization at the sentence-level \citep{Zhang2016ArgRewrite:Writings}. 
Our new system, ArgRewrite V.2, augmented the prior work by developing two additional interfaces for \textit{binary sentential} (Interface B) and \textit{fine-grained subsentential} (Interface D) revision categorization. Impact of different interfaces on students' writings are evaluated using both survey and writing improvement data \citep{Afrin2021EffectiveWriting}. 




\section{Conclusion}
We have introduced ArgRewrite V.2, a corpus of revisions that are collected from argumentative essays written by university students in response to a writing prompt, and revised in response to some revision feedback.  
Revisions are semi-automatically aligned at both sentential and subsentential units, and each revision unit, then, manually annotated by domain experts with its coarse and fine-grained purpose category.

Aside from the annotated revisions, ArgRewrite V.2 also includes additional meta-data such as participants' demographic and self-regulation survey, as well as evaluative feedback on the drafts. 
To demonstrate the potential of ArgRewrite as a resource for revision analysis and other NLP applications, we explored usages of the corpus in a variety of automatic revision purpose prediction tasks.

\bibliographystyle{spbasic}      
\bibliography{references.bib,bibliography.bib}   

\newpage
\appendix


\section{Writing Prompt}
\label{apx:prompt}

Students are asked to read a brief article about self-driving cars, and then write a short argumentative essay in response to the following prompt: 

\begin{quote}
\itshape
In this argumentative writing task, imagine that you are writing an op-ed piece for the Pittsburgh City Paper about self-driving cars. The editor of the paper has asked potential writers, like you, to gather information about the use of self-driving cars, and argue whether they are beneficial or not beneficial to society.

In your writing, first, briefly explain both the advantages and disadvantages of self-driving cars. Then, you will choose a side, and construct an argument in support of self-driving cars as beneficial to society, or against self-driving cars as not beneficial to society. A high quality op-ed piece maintains a clear position on the issue and uses supporting ideas, strong evidence from the reading, explanations of your ideas and evidence, and a counter-argument. Furthermore, a high quality op-ed piece is clearly organized, uses precise word choices, and is grammatically correct.
\end{quote}

\section{Annotation Guideline}
\label{apx:guideline}
\subsection{Alignment annotation}
The essays for each draft are tokenized into sentences. The sentences are enumerated from 1
to N according to their occurrence in the essay as `Sentence Index'. For `Aligned Index', each
sentence in the revised draft is assigned the index of its aligned sentence in the original draft.
Also, each sentence in the original draft is assigned the index of the aligned sentence in the
revised draft. If a sentence is newly added, it will be marked as ADD. If a sentence is deleted
from the old draft, it will be marked as DELETE.

\subsubsection{Rules}
\begin{enumerate}
    \item Every sentence should either be aligned (one-to-one, one-to-many, many-to-one) or
marked as ADD or DELETE. Only the alignment from the Old Draft to New Draft contains
``DELETE'' and only alignment from the New Draft to Old Draft contains ``ADD''.

    \item For one-to-one case, align the sentences if the revised sentence is either replication or
modification of the original sentence with one or several of the following changes:
\begin{enumerate}
    \item Addition/deletion of some content within the sentence
    \item Modification of words, phrases
    \item Restatement of the ideas of the sentence
\end{enumerate}

The aligned sentences should be either syntactically or semantically close and within the
same/similar context (i.e. the paragraphs the sentences belong to should be similar)
\begin{itemize}
    \item Syntactically similar: The two sentences look explicitly similar to each other. (i.e. the difference between the two sentences should be a small ratio of the whole sentence. For example, a sentence with less than 10 words should have at most 2 words that are different (Does not count the change of words in the same stem, e.g. change-> changes)).
    \item Semantically similar: The two sentences describe the same information, or the revised sentence adds/deletes information on the basis of the original sentence
\end{itemize}
    \item For many-to-one and one-to-many cases, only align when multiple sentences are
syntactically similar to some part of the one target sentence. When multiple sentences can
be combined without major addition/deletion/modification of words/phrases to construct
the aligned sentence. Or, when one sentence can be divided to construct the aligned
sentences without major addition/deletion/modification of words/phrases. It should also
be explicit and better to align the target sentence to the group of sentences than to align
the target sentence to one or some of the sentences.
\end{enumerate}

\subsection{Revision Purpose Annotation}
\begin{figure}[h]
    \centering
    \includegraphics[width=0.99\linewidth]{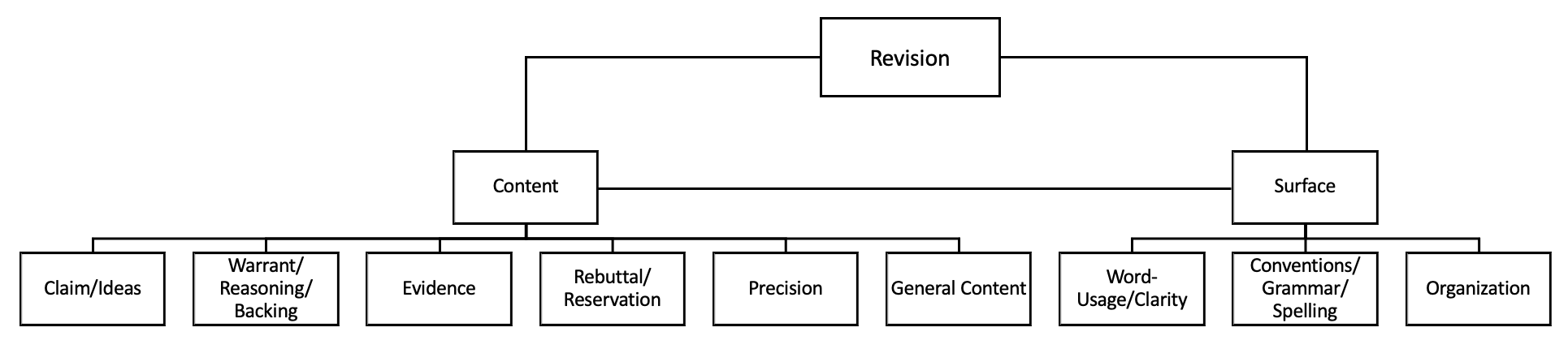}
    \caption{Revision Purpose Schema.}
    \label{fig-apx:rev purpose}
\end{figure}

Each aligned sentence (including ADD and DELETE) should have ONE major revision purpose. As shown in the Figure~\ref{fig-apx:rev purpose}, each revision purpose can be classified as two higher-level changes – surface and text/content. These change can be further categorized into 9 major revision purposes. The annotator is required to annotate ONLY the major revision purpose. Annotator has to obey the following rules to decide the major revision purpose type.

\subsubsection{Rules} 
\begin{enumerate}
    \item Importance Orders of Revision Purposes (Higher to lower): 

 
The importance of different revision purpose type is different, when there are multiple revision purpose types in one revised sentence, make sure that the more important one is selected. The following sub-rules explains more specific details for cases where the decision of the appropriate revision purpose can be difficult.

\begin{itemize}
    \item Claim/Ideas vs. Warrant/Reasoning/Backing
    
An essay can have one major claim and several sub-claims to support the major claim. These sub-claims are usually in the form of reasoning to support the major claim. Thus the differentiation of sub-claim and reasoning for the major claim can be ambiguous. We ask the annotators to think of the Claim and Reasoning as a hierarchical tree structure. The leaves of the tree are marked as ``Warrant/Reasoning/Backing'' while the others are marked as ``Claim/Ideas''. Hence, there should be no ``Warrant/Reasoning/Backing'' without a “Claim/Ideas” seen before. In specific, if the major idea of the essay is further supported or objected by other sentences, it is considered as a Claim. If the sentence cannot be classified as Evidence/Rebuttal of the Claim, but the sentence contains elements backing or reasoning for or against the Claim, it should be annotated as ``Warrant/Reasoning/Backing''.

    \item General Content vs. Warrant/Reasoning/Backing
    
Differentiating General Content and Reasoning can be difficult as they both often occur after the author proposes a claim. To differentiate the two categories, the annotator is required to distinguish whether the author is suggesting his position for his claim in the sentences or not. If the annotator senses the author's sentiment position towards his claim, then it should be ``Warrant/Reasoning/Backing'', whereas it should be ``General Content''.

    \item Evidence vs. Warrant/Reasoning/Backing
    
These two categories are similar as they both provide support to the authors' claim. The annotators are required to distinguish these two categories according to whether the sentences are stating facts. The facts can be (1) Citation: the citation of papers, reports, news and books. (2) Example: facts of history or personal experiences. (3) Scientific proof. If there are facts involved, it is marked as Evidence, otherwise it is marked as Warrant.

    \item Conventions/Grammar/Spelling vs. Word Usage/Clarity
    
These two genres are similar as they do not change the content of the text and improve the quality of the text. The annotators are required to make the judgment according to the question: Are there spelling/grammar mistakes in the original draft and has this mistake been addressed in the new draft? If the ONLY a mistake is addressed, it should be marked as ``Grammar/Spelling''.

\item Precision vs Word-Usage/Clarity
    
These two genres are not similar but annotating can be confusing. When there is a word/phrase change in the sentence that significantly change the specificity level of a sentence to make it more specific or general as a content revision, then it is a precision change, otherwise it will be a word-usage change. 

    \item Claim/Idea vs Word-Usage/Clarity
    
These two genres are not similar but annotating can be confusing when there is a word/phrase change in the claim of the essay affecting major claim. If the change of the sentence affects/changes the claim of the essay, it should be annotated as Claim/Ideas instead of Word-Usage/Clarity. Because a change in the claim affects the subsequent changes of warrant/evidence. A feedback of the revision as claim change would help writers understand and think about the changes of the essay better than a feedback of a word usage.

    \item Organization vs General Content Development
    
Although these two categories seem very different, annotators need to be very careful while annotating these two. General content changes are usually heavy changes in the sentence (compared to Word-Usage) or added and deleted sentences. If merged or split sentences do not have major change in words, it should be Organization. However, if those sentence have major change in words so that it is better to consider them as individual sentence rather than aligned sentence, then it should be annotated as General Content. Sometimes reordered sentences maybe aligned as DELETE and then ADD. In those cases, it should be considered as Organization rather than DELETE General Content and then ADD General Content.
\end{itemize}

    \item Focus on WHAT than WHERE 

It is not necessarily that revisions made on the thesis of the paragraph are Claim/Idea changes, the type of the change should be determined according to what the author really has changed. For example, in a Claim sentence of a paragraph, if the author added a clause in the new sentence for reasoning the claim, the change would be a Warrant/Reasoning/Backing change; if the author only replaced some word with a more appropriate form of word, the annotator should mark it as Word usage change. However, if the change affects the claim it should be a Claim/Ideas change as stated before.

    \item Read and understand the prompt before the annotation. 

Sometimes the annotation of revision purpose could be different according to what the author is really targeting. So it is critically important that the annotator read and understand the prompt before the annotation. For example, in a regular essay, a sentence change from “Fidel Castro would be a good example for this case” to “Saddam Hussein would be a good example for this case” would typically be “Evidence”. However, if the prompt of the essay writing assignment is “Put the contemporaries at different levels of Hell”, then the annotation would be “Claim/Ideas”. 

We have developed an annotation tool to ease the annotation of alignments, the tool automatically breaks the text to sentences and the annotator only needs to do the annotation on the interface. After the alignment completes, the annotator can select the type of the revision purpose. Check out more details of the tool in the annotation too manual.
\end{enumerate}

\clearpage
\newpage

\section{Interfaces}
\label{apx:interfaces}

Each participant was randomly assigned to one of the following four interfaces, which provided different types of feedback on differences between Draft1 and Draft2, including the size of the revision unit span and the granularity of the revision purpose category. 
For more details refer to \citep{Afrin2021EffectiveWriting}.

\begin{figure}[h]
  \centering
  \begin{subfigure}{0.48\linewidth}
        \fbox{\includegraphics[width=0.95\linewidth]{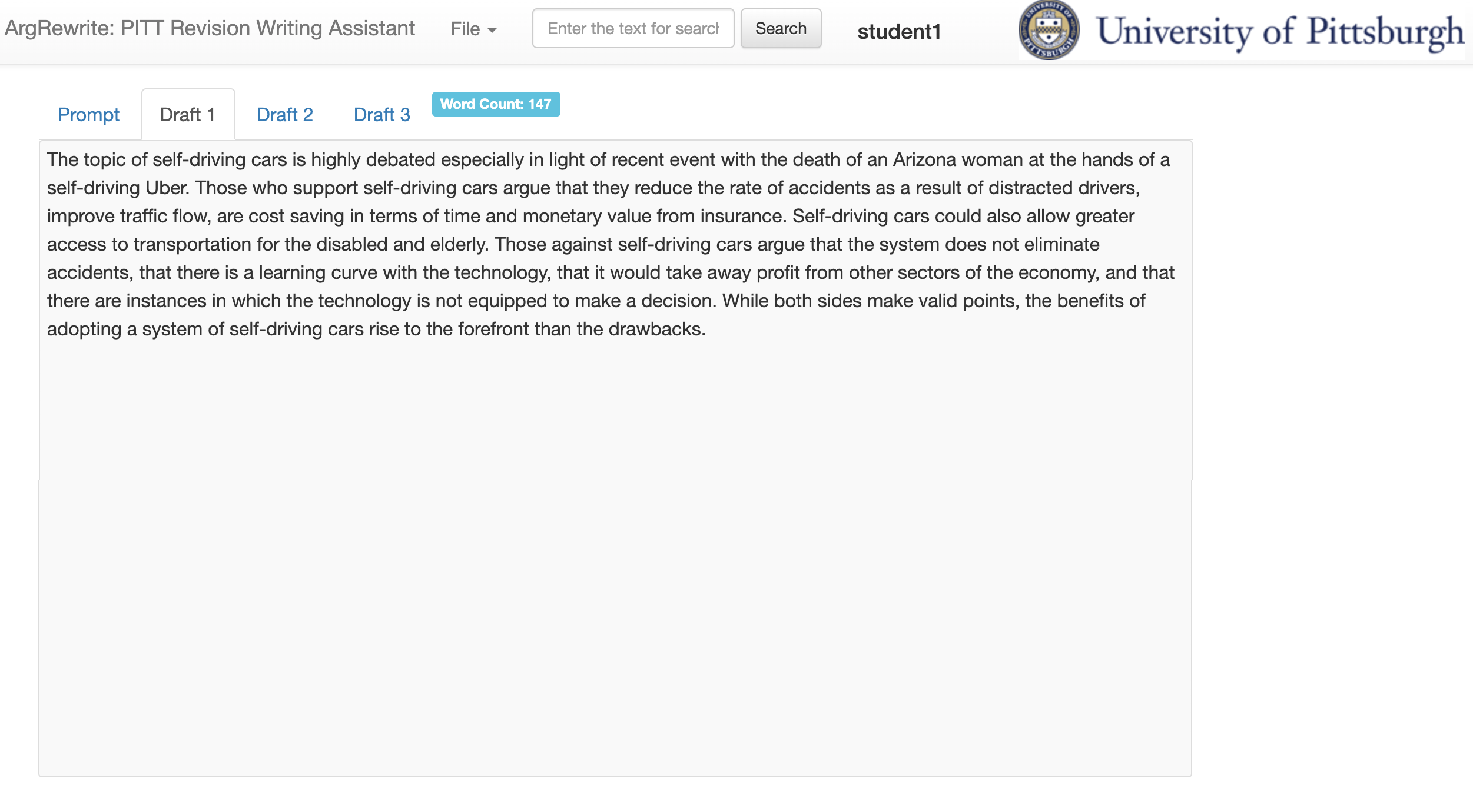}}
        \caption{Interface A}
        \label{fig:intA}
    \end{subfigure}
    ~
   \begin{subfigure}{0.48\linewidth}
        \fbox{\includegraphics[width=0.95\linewidth]{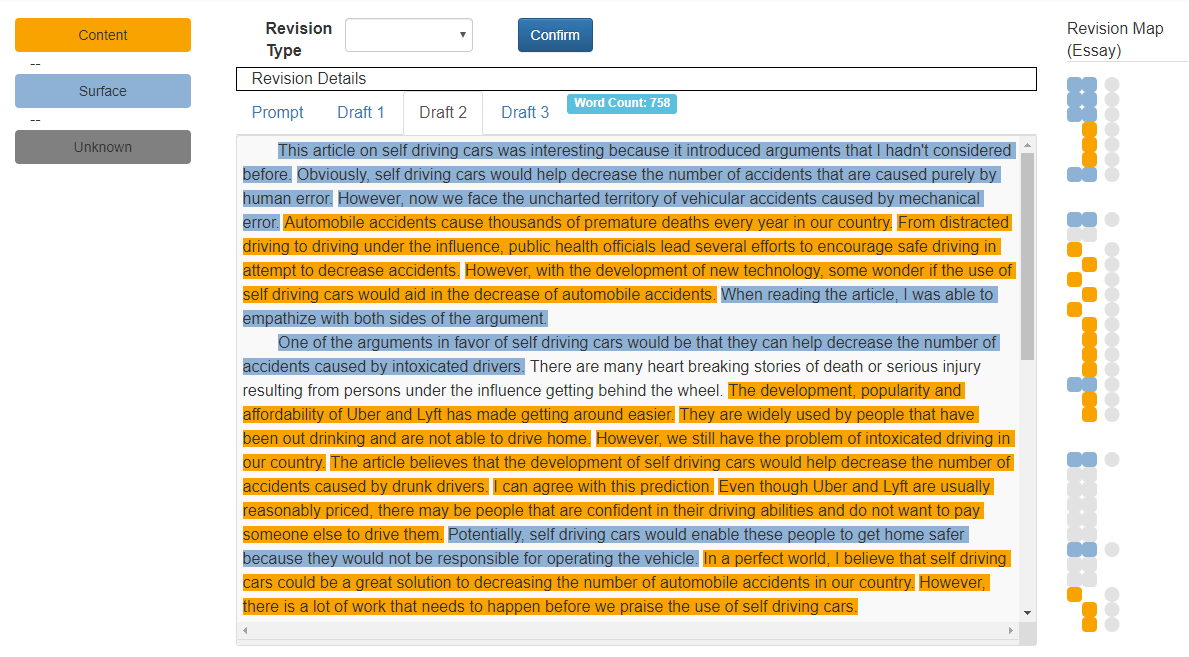}}
        \caption{Interface B}
        \label{fig:intB}
    \end{subfigure}
    \begin{subfigure}{0.48\linewidth}
        \fbox{\includegraphics[width=0.95\linewidth]{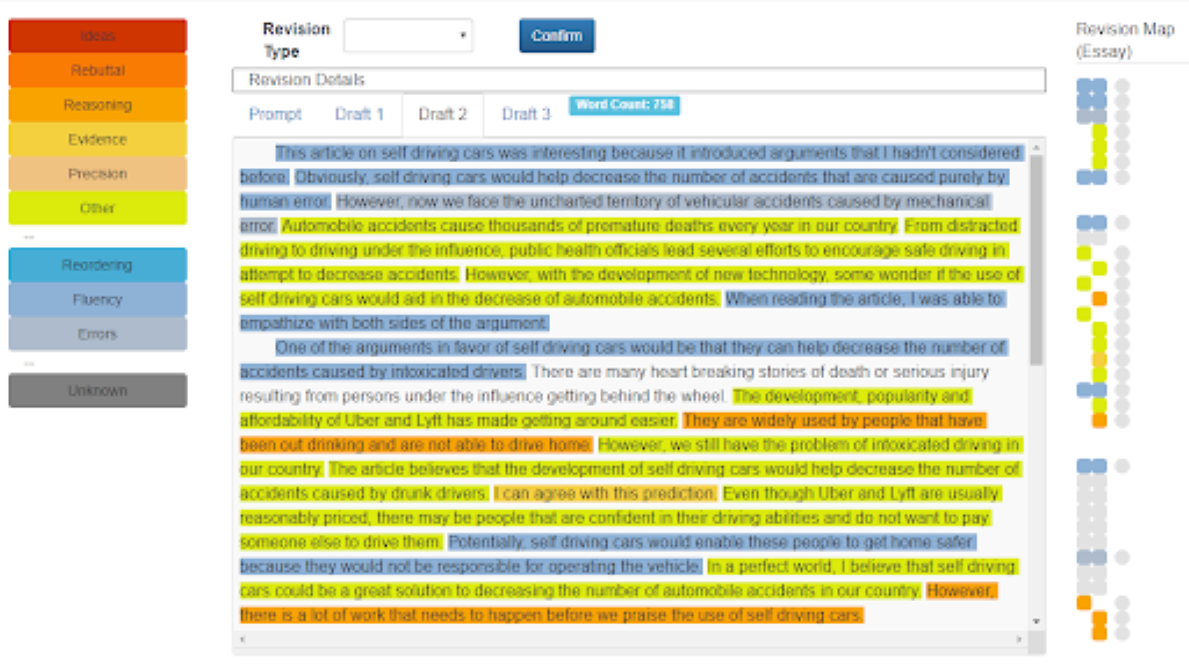}}
        \caption{Interface C}
        \label{fig:intC}
    \end{subfigure}
    ~
    \begin{subfigure}{0.48\linewidth}
        \fbox{\includegraphics[width=0.95\linewidth]{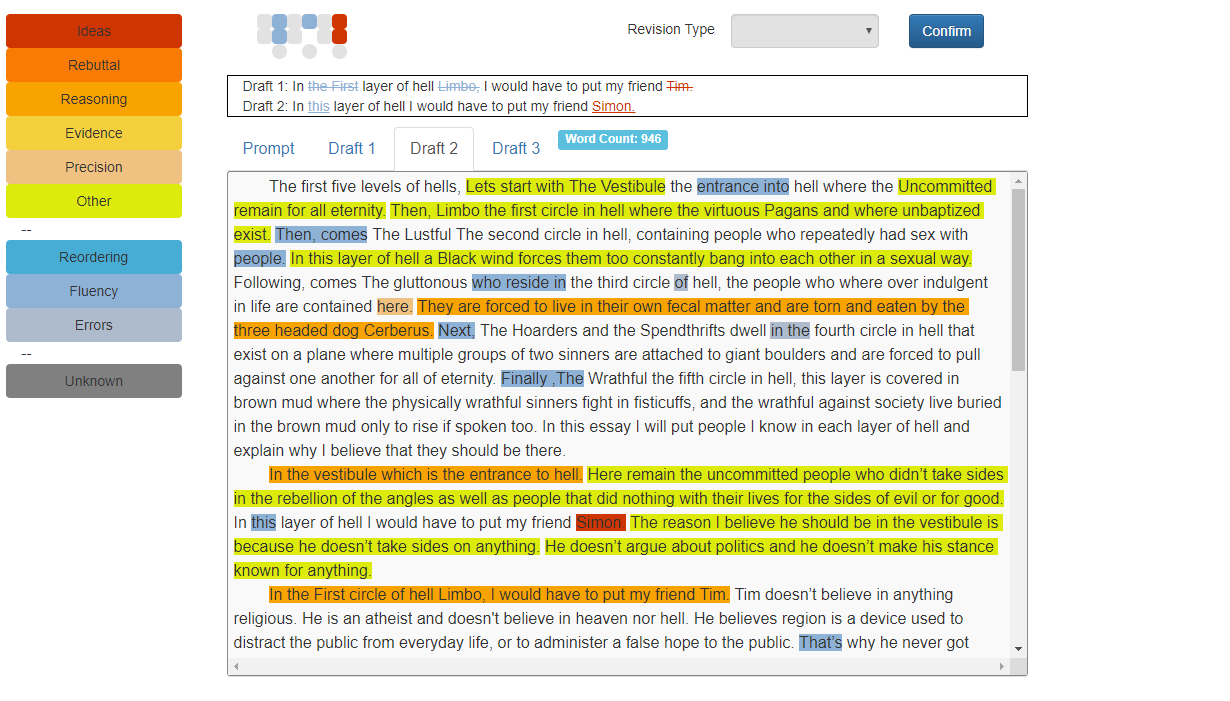}}
        \caption{Interface D}
        \label{fig:intD}
    \end{subfigure}
  \caption{Screenshot of Different Conditions, where warmer colors indicate content revisions and colder colors indicate surface revisions. (a) No Feedback; (b) Sentence-Level feedback with coarse-grained (surface vs. Content) revision purposes; (c) Sentence-Level feedback with fine-grained revision purposes; (d) Subsentential-Level feedback with fine-grained revision purposes }
  \label{fig:int}
\end{figure}

\begin{itemize}
    \item \textbf{Interface A.} The 20 participants assigned to this condition were shown only the changed sentences without any further feedback (Figure~\ref{fig:intA}); 
    
    \item \textbf{Interface B.} The 22 participants assigned to this condition were shown sentence-level differences, as either a surface or content revision (Figure~\ref{fig:intB}); 
    
    \item \textbf{Interface C.} The 22 participants assigned to this condition were shown sentence-level differences with fine-grained revision purposes (Figure~\ref{fig:intC}); 

    \item \textbf{Interface D.} The 22 participants assigned to this condition were shown subsentential differences with fine-grained revision purposes (Figure~\ref{fig:intD})
\end{itemize}

\clearpage
\newpage

\section{Feedback}
\label{apx:feedback}

\subsection{Prior Study's Feedback}
\label{apx:feedback:v1}

The same feedback given to all students in the prior version of the corpus \citep{Zhang2015AnnotationRevisions}:

\begin{quote}
\textit{Strengthen the essay by adding one more example or reasoning for the claim; then add a rebuttal to an opposing idea; keep the essay at 400 words. }
\end{quote}

\subsection{Personalized Feedback Example}
\label{apx:feedback:example}

An example of a personalized feedback message:

\begin{quote}
\textit{Thank you for your participation in the study. Your draft has been read, and feedback from an expert writing instructor is written below. We advise that you use this feedback when you revise.}

\textit{The strengths of your essay include:}
\begin{itemize}
    \item \textit{All claims have relevant supporting evidence, though that evidence may be brief or general.}
    \item \textit{You respond to one, but not all parts of the prompt. However, your entire essay is focused on the prompt.}
\end{itemize}
\textit{Areas to improve in your essay include:}  
\begin{itemize}
    \item \textit{You provided a statement that somewhat show your stance for or against self-driving cars, but it is unclear, or is just a restatement of the prompt.}
    \item \textit{Your essay’s sequence of ideas is inconsistent, with some clear and some unclear progression. }
    \item \textit{Your essay does not include a rebuttal.}
\end{itemize}

\end{quote}

\subsection{Scoring Rubric}
\label{apx:feedback:rubric}

Each participants is given a personalized feedback in the form of lists of 2-4 strengths and 2-4 weaknesses that characterized their first draft of the essay based on the following scoring rubric: 

\setlength\tabcolsep{2.9pt} 
\begin{longtable}{m{.6cm}|m{2.4cm}|m{2.7cm}|m{2.45cm}|m{2.7cm}}
    \caption{Argumentative Essay Rubric }\\

     &   \textbf{1-Poor} & \textbf{2-Developing} & \textbf{3-Proficient} & \textbf{4-Excellent} \\ 
     \hline
     \endfirsthead
    &   \textbf{1-Poor} & \textbf{2-Developing} & \textbf{3-Proficient} & \textbf{4-Excellent} \\ 
    \hline
     \endhead
     \rotatebox{90}{\parbox{1.8cm}{\bf Response to \\ prompt}} & 
         The essay is off topic, and does not consider or respond to the prompt in any way. &
         The essay addresses the topic, but the entire essay is not focused on the prompt. The author may get off topic at points. &
         The author responds to one, but not all parts of the prompt, but the entire essay is focused on the prompt. &
         The author responds to all parts of the prompt and the entire essay is focused on the prompt. \\ \hline
         
     \rotatebox{90}{\bf Thesis} &
         The author did not include a statement that clearly showed the author’s stance for or against self-driving cars. &
         The author provided a statement that somewhat showed the author’s stance for or against self-driving cars, though it may be unclear or only a restatement of the essay prompt. &
         The author provided a brief statement that reflects a thesis, and is indicative of the stance the author is taking toward self-driving cars. &
         The author provided a clear, nuanced and original statement that acted as a specific stance for or against self-driving cars. \\  \hline
         
     \rotatebox{90}{\bf Claims} & 
        The author’s claims are difficult to understand or locate. & 
        The author’s claims are present, but are unclear, not fully connected to the thesis or the reading, or the author makes only one claim multiple times. & 
        The author makes multiple, distinct, and clear claims that align with either their thesis or the given reading, but not both. & 
        The author makes multiple, distinct claims that are clear, and align with both their thesis statement and the given reading. They fully support the author’s argument. \\  \hline
        
     \rotatebox{90}{\bf Evidence for Claims} & 
        The author does not provide any evidence to support thesis/claims. & 
        Less than half of claims are supported with relevant or credible evidence or the connections between the evidence and the thesis/claims is not clear. & 
        All claims have relevant supporting evidence, though that evidence may be brief or general. The source of the evidence is credible and acknowledged/cited where appropriate. & 
        The author provides specific and convincing evidence for each claim, and most evidence is given through detailed personal examples, relevant direct quotations, or detailed examples from the provided reading. The source of the evidence is credible and acknowledged/cited where appropriate. \\  \hline
        
     \rotatebox{90}{\bf Reasoning} & 
        The author provides no reasoning for any of their claims. & 
        Less than half of claims are supported with reasoning or the reasoning is so brief, it essentially repeats the claim. Some reasoning may not appear logical or clear. & 
        All claims are supported with reasoning that connect the evidence to the claim, though some may not be fully explained or difficult to follow. & 
        All claims are supported with clear reasoning that shows thoughtful, elaborated analysis. \\  \hline
    
    \rotatebox{90}{\parbox{2cm}{\bf Reordering / \\Organization}} &
    	The sequence of ideas/claims is difficult to follow and the essay does not have an introduction, conclusion, and body paragraphs that are organized clearly around distinct claims. &
    	The essay’s sequence of ideas is inconsistent, with some clear and some unclear progression of ideas OR the essay is missing a distinct introduction OR conclusion. &
    	The essay has a clear introduction, body, and conclusion and a logical sequence of ideas, but each claim is not located in its own separate paragraph. &
    	The essay has an introduction, body and conclusion and a logical sequence of ideas. Each paragraph makes a distinct claim. \\ \hline
      
    \rotatebox{90}{\bf Rebuttal} &
    	The essay does not include a rebuttal. &
    	The essay includes a rebuttal in the sense that it acknowledges another point of view, but does not explore possible reasons why this other viewpoint exists. &
    	The essay includes a rebuttal in the form of an acknowledgement of a different point of view and reasons for that view, but does not explain why those reasons are incorrect or unconvincing. &
    	The essay explains a different point of view and elaborates why it is not convincing or correct. \\ \hline
    	
    \rotatebox{90}{\bf Precision} &
    	Throughout the essay, word choices are overly informal and general (e.g., “I don’t like self-driving cars because they have problems.”). &
    	Word choices are mostly overly general and informal, though at times they are specific. &
    	Word choices are mostly specific though there may be a few word choices that make the meaning of the sentence vague. &
    	Throughout the essay, word choices are specific and convey precise meanings (e.g., “Self-driving cars are dangerous because the technology is still not advanced enough to address the ethical decisions drivers must make.”) \\ \hline
    
    \rotatebox{90}{\bf Fluency} &
    	A majority of sentences are difficult to understand because of incorrect/ inappropriate word choices and sentence structure. &
    	A noticeable number of sentences are difficult to understand because of incorrect/ inappropriate word choices and sentence structure, although the author’s overall point is understandable. & 
    	Most sentences are clear because of correct and appropriate word choices and sentence structure. &
    	All sentences are clear because of correct and appropriate word choices and sentence structure. \\ \hline
    	
    \rotatebox{90}{\parbox{2.3cm}{ Conventions/Gr-\\ammar/Spelling}} &
    	The author makes many grammatical or spelling errors throughout their piece that interfere with the meaning. &
    	The author makes many grammatical or spelling errors throughout their piece, though the errors rarely interfere with meaning. &
    	The author makes few grammatical or spelling errors throughout their piece, and the errors do not interfere with meaning. &
    	The author makes few or no grammatical or spelling errors throughout their piece, and the meaning is clear. \\
    \hline 

    \label{tab:rubric}

\end{longtable}
\setlength\tabcolsep{6pt} 

\newpage

\section{Pre-Study Questionnaires}
\label{apx:pre-study}
\begin{itemize}
    \item Are you an undergraduate or graduate student?
    \item What is your current year of study?
    \item Is English your native language?
    \item What is your native language?
    \item When writing an essay/paper for a class, how many drafts (that are not required by the class) do you typically write?
    \item Overall, how confident are you with your writing?
    \item Please tell us how comfortable you feel about writing in the English language versus writing in your primary language.
    \item What aspects of writing do you think you are good at?[Click all that apply]
    \item What aspects of writing do you think you can improve?[Click all that apply]
    \item I typically set aside routine, planned times to complete writing tasks.
    \item I typically create an outline of my writing before I begin any writing task.
    \item I typically seek out feedback from others on my writing.
    \item I typically plan time for multiple revisions of my writing.
    \item I typically set revision goals for myself to meet the requirements of a writing task.
    \item The revision goals I set for myself focus mostly on developing the content or thesis.
    \item The revision goals I set for myself focus mostly on surface level changes (e.g. grammar, spelling, organization and word clarity).
    \item While I am revising, I typically look back at or think about my previous draft(s) to refine my essay.
    \item While I am revising, I typically look back at or think about feedback from others to refine my essay.
    \item While I am revising, I typically think about the reader's expectations.
    \item While I am revising, I typically address grammatical errors.
    \item While I am revising, I typically try to develop the content or thesis.
    \item When I make a revision, I reread the sentence, paragraph, or whole essay to see whether my revision improved the essay.
    \item I can meet the requirements of a writing task without revising.
    \item I am confident in my writing and revising abilities.
\end{itemize}

\newpage

\section{Post-Study Questionnaires}
\label{apx:post-study}

Followings are the questions asked from all students, regardless of the interface they assigned to: 

\begin{itemize}
    \item The system allows me to have a better understanding of my previous revision efforts.
    \item I find the system easy to use.
    \item My interaction with the system is clear and understandable.
    \item The system helps me to recognize the weakness of my essay.
    \item The system encourages me to make more revisions (quantity) than I usually do.
    \item The system encourages me to make more meaningful revisions (quality) than I usually do.
    \item Overall the system is helpful to my writing.
    \item I put a lot of effort into writing and revising this essay.
    \item How could the system be more helpful?
\end{itemize}

\noindent Following questions are only asked form the students who were assigned to \textbf{Interface A}:
\begin{itemize}
    \item What led you to notice that some parts of your essay needed to be revised?
    \item Was this revision process similar to how you normally revise your essays?
\end{itemize}

\noindent Following questions are only asked form the students who were assigned to \textbf{Interface B}:
\begin{itemize}

    \item I found the overview page to be useful.
    \item The description of the purpose of my revisions inspired me to make more revisions. 
    \item I  found it useful to see my revision purposes highlighted in different colors (ie. Warm and cold colors)
    \item I found the revision map visualization useful. 
    \item I found the small window of revision details to be useful.
    \item In general, I found it helpful to know whether my revision was a surface or content level change.
    \item My revision purposes were most often indicated correctly by the system.
    \item I trust the feedback that the system gave me.
    \item What influenced your decision to make revisions to Draft3?
\end{itemize}

\noindent In addition to all the question that are asked from the students in Interface B, students who were assigned to \textbf{Interface C} are also asked the following question:

\begin{itemize}
    \item I found it helpful to have the specific purposes of my revisions indicated (e.g. claim, evidence, warrant, etc.).
\end{itemize}

\noindent In addition to all the question that are asked from the students in Interface C, students who were assigned to \textbf{Interface D} are also asked the following question:

\begin{itemize}
    \item The system accurately highlighted each, specific area of text that I revised (this area of text could be as small as a word, or as large as a sentence).
\end{itemize}

\newpage

\section{Term Frequency Representation}
\label{apx:term-freq}

The term frequency representation is a vector with the size of the total number of classes.
The \texttt{spaCy} library recognizes 19 different POS tags, so the term frequency representation of the POS is an array with length 19, where each index represents a POS tag, and the number at each index represents the total number of words in a sentence that has that POS tag.
    
For example, consider the following sentences and the associated POS of its word:

\begin{table}[h]
    \centering
    \begin{tabular}{lllll}
         \underline{this} & \underline{is}  & \underline{a} & \underline{revised}  & \underline{sentence}\\
         DET              & VERB            & DET           & VERB                 & NOUN     \\
         & 
    \end{tabular}
    \label{tab:tf-example}
\end{table}

The POS term frequency representation of this sentence would be $[0000002010000000200]$, where each index represent the number of words with one of the 19 POS tags, for example, 
the number at index 0 represents the number of ADJECTIVEs (there is none of the in the sentence so the value is 0), 
the 6th index represents the number of DETs (there are two words with this tag so the value is 2), 
the 8th index represents NOUNs (there is one word with this tag so the value is 1), 
and the 16th index represents the number of VERBs (we have two verbs so the values is 2).

\section{Transition Words}
\label{apx:trans-words}

Table~\ref{tab:apx:trans_words} includes the list of words we used for calculating the term frequency representation of transition words as a feature for revision purpose classification tasks.
We collected these words from multiple transition word lists published by the writing centers of some universities\footnote{\texttt{https://writing.wisc.edu/handbook/style/transitions/} \\ \texttt{http://writing2.richmond.edu/writing/wweb/trans1.html} \\ \texttt{https://writingcenter.unc.edu/tips-and-tools/transitions/}} and filtered for those words and categories that we though might correspond to our revision purpose categories.

\begin{table}[h]
    \centering
    \caption{Transition Words used for Training Revision Purpose Classifiers}
    \begin{tabular}{|r|p{9.35cm}|}

         \hline 

         \textbf{Group} & \textbf{Words} \\
         \hline \hline
         \textbf{Reasoning} & consequently, clearly, then, furthermore, additionally, moreover, because, besides, also\\
         
         \textbf{Evidence} & (as an) illustration, e.g., (for) example, (for) instance, specifically, (to) demonstrate, (to) illustrate \\
         
         \textbf{Rebuttal} & however, but, yet, although, despite, (in) contrast, nevertheless, nonetheless, notwithstanding, (on the) contrary, otherwise, though, yet\\
         
         \textbf{Conclusion} & therefore, hence, conclusion, consideration, indeed, finally, lastly\\
         
         \textbf{Details} & Specifically, especially, (in) particular, (to) explain, (to) list, (to) enumerate, (in) detail, namely, including \\
            
        \textbf{Causation} & accordingly, so, because, consequently, hence, since, therefore, thus \\
        
         \hline 
    \end{tabular}
    \label{tab:apx:trans_words}
\end{table}

\end{document}